\documentclass[accepted]{uai2023} %

\usepackage[american]{babel}
\usepackage{natbib} %
\usepackage{mathtools} %
\usepackage{booktabs} %
\usepackage{tikz} %

\usepackage{microtype}
\usepackage{graphicx}
\usepackage{subfigure}
\usepackage{booktabs} %
\usepackage{xcolor}
\usepackage{multirow}
\usepackage{array}
\usepackage{tabularx}
\usepackage{wrapfig}
\usepackage{caption}

\usepackage{hyperref}
\hypersetup{colorlinks,linkcolor={teal},citecolor={magenta},urlcolor={red}}

\usepackage{hyperref}
\usepackage{amsmath}
\usepackage{mathrsfs}
\usepackage{bm}
\usepackage{amssymb}
\DeclareUnicodeCharacter{2212}{-}
\usepackage[ruled,vlined]{algorithm2e}

\title{ \texttt{USIM-DAL}: Uncertainty-aware Statistical Image Modeling-based Dense Active Learning for Super-resolution}

\author[1]{Vikrant Rangnekar$^{*}$}
\author[2]{Uddeshya Upadhyay$^{*}$}
\author[2,3]{Zeynep Akata}
\author[1]{Biplab Banerjee}
\affil[1]{%
    Centre for Machine Intelligence and Data Science (CMInDS), 
    IIT Bombay
}
\affil[2]{%
    University of Tübingen
}
\affil[3]{%
    Max Planck Institute for Intelligent Systems, Tübingen
}

\begin{document}
\maketitle

\begin{abstract}
    Dense regression is a widely used approach in computer vision for tasks such as image super-resolution, enhancement, depth estimation, etc. However, the high cost of annotation and labeling makes it challenging to achieve accurate results. We propose incorporating active learning into dense regression models to address this problem. Active learning allows models to select the most informative samples for labeling, reducing the overall annotation cost while improving performance. Despite its potential, active learning has not been widely explored in high-dimensional computer vision regression tasks like super-resolution. We address this research gap and
    propose a new framework called \textit{USIM-DAL} that leverages the statistical properties of colour images to learn informative priors using probabilistic deep neural networks that model the heteroscedastic predictive distribution allowing uncertainty quantification. Moreover, the aleatoric uncertainty from the network serves as a proxy for error that is used for active learning. Our experiments on a wide variety of datasets spanning applications in natural images (visual genome, BSD100), medical imaging (histopathology slides), and remote sensing (satellite images) demonstrate the efficacy of the newly proposed \textit{USIM-DAL} and superiority over several dense regression active learning methods.
\end{abstract}

\section{Introduction}
\label{sec:intro}
\vspace{-5pt}
\let\thefootnote\relax\footnotetext{$^*$Both the authors contributed equally}
The paradigm of dense prediction is very important in computer vision, given that pixel-level regression tasks like super-resolution, restoration, depth estimation etc., help in holistic scene understanding. 
A common example of a pixel-level (i.e., dense) regression task is \textit{Image super-resolution} (SR) is the process of recovering high-resolution (HR) images from their low-resolution (LR) versions. It is an important class of image processing techniques in computer vision, deep learning, and image processing and offers a wide range of real-world applications, such as medical imaging~\citep{li2021review}, satellite imaging~\citep{verpoorter2014global}, surveillance~\citep{1220866} and security~\citep{7518116}, and remote sensing~\citep{yang2015remote}, to name a few.
The well-performing techniques for super-resolution often rely on deep learning-based methods that are trained in a supervised fashion, requiring high-resolution data as groundtruth. However, the acquisition of high-resolution imaging data (to be served as labels) for many real-world applications may be infeasible.
Consider the example of histopathology microscopy from medical imaging, where the typical digital microscope takes significantly longer to acquire the high-resolution scans (i.e., at high magnification) image of the slide than low-magnification~\citep{aeffner2018digital,hamilton2014digital}. Moreover, the acquired high-resolution scans also have a significantly larger memory footprint leading to an increase in storage resources~\citep{bertram2017pathologist}.
Similarly, acquiring high spatial resolution images from satellites for remote sensing requires expensive sensors and hardware and has significantly higher operating costs~\citep{cornebise2018witnessing,cornebise2022open}. In such scenarios, generating a large volume of training samples is infeasible.

\begin{figure*}[!t]
    \centering
    \includegraphics[width=\textwidth]{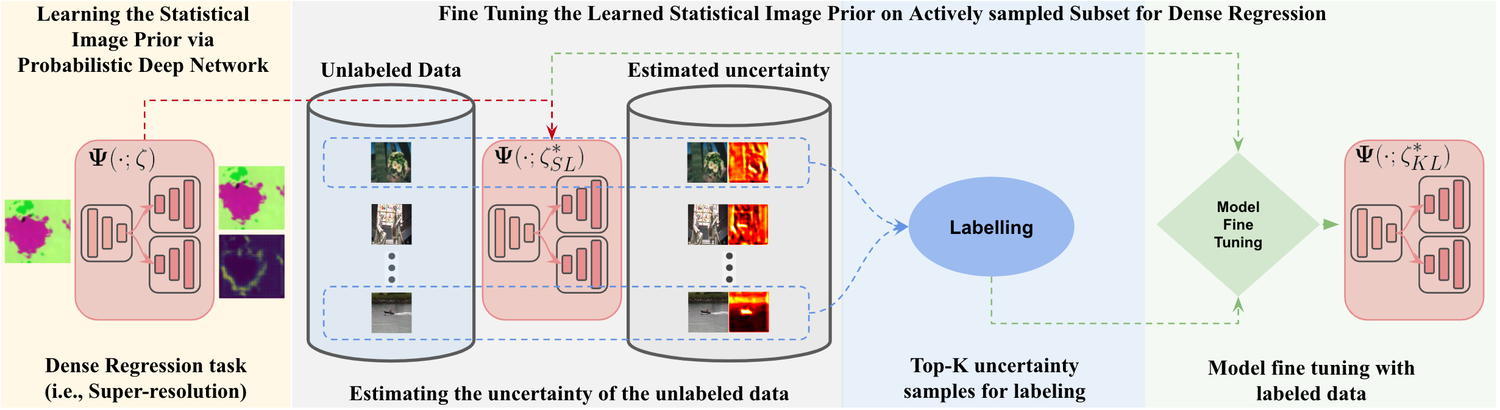}
    \vspace{-10pt}
    \caption{The proposed framework \textit{USIM-DAL}. (Left-to-right) We train a probabilistic deep network for a dense regression task (e.g., super-resolution) on synthetic samples obtained from statistical image models as described in Section~\ref{sec:method}. The pre-trained model is used to identify the high-uncertainty samples from the domain-specific unlabeled set. Top-K highly uncertain samples are chosen for labeling on which the pre-trained network is further fine-tuned.
    }
    \label{fig:arch}
    \vspace{-10pt}
\end{figure*}

As a remedy, concepts like zero-shot SR or single-image SR have been proposed. Nevertheless, zero-shot SR still requires ample supervision from the test image patches~\citep{shocher2018zero} to learn the transferrable model for novel scenarios with divergent distributions~\citep{soh2020meta}, and the performance of the single-image SR models is still affected by the lack of sufficient labeled data~\citep{Lim_2017_CVPR_Workshops}. Notwithstanding these discussions, there are situations where there are restrictions on dealing with training samples within a pre-defined budget. 
For example, in histopathology microscopy, the constraint on available resources may allow high-resolution acquisition for only a limited number of patients/microscopy slides.
One of the viable solutions in this regard is to select a subset of highly representative training samples from the available training set while respecting the budget and deploying them to train the SR model. This corresponds to the notion of active learning for subset selection. However, selecting the subset is challenging considering the fact that we need a quantitative measurement for the eligibility of a given training LR-HR pair to be selected.
Many works have explored different \textit{query functions} to select a subset to label from a larger dataset~\citep{Beluch_2018_CVPR,gorriz2017cost,roy2001toward}. However, most of them have been applied to classification or low-dimensional regression problems~\citep{jain2016active}, and there still exists a gap on how to address this for dense regression tasks (e.g., super-resolution).
Active learning technique to label those points for which the current model is least certain has been studied well in the context of classification~\citep{yang2015multi}. While there are recent advances in uncertainty estimation using neural networks for dense regression~\citep{NIPS2017_2650d608,upadhyay2022bayescap}, it is yet to be studied if they can be leveraged in active learning for dense regression.

In summary, our contributions are as follows:
(i)~We show how statistical image models can help alleviate the need for a large volume of high-resolution imaging data.
(ii)~We show that probabilistic deep networks, along with the statistical image models, can be used to learn informative prior about niche domain datasets that may allow limited access to high-resolution data. 
(iii)~Our probabilistic deep network trained with the statistical image models allows us to estimate the uncertainty for the sample in a niche domain that can be leveraged for active learning as illustrated in Figure~\ref{fig:arch}.

\vspace{-5pt}
\section{Related Work}
\label{sec:rw}
\vspace{-5pt}
\paragraph{Active Learning.} These are a set of techniques that involve selecting a minimal data subset to be annotated, representing the entire dataset, and providing maximum performance gains. Querying strategies for active learning can be broadly categorized into three categories: heterogeneity-based, performance-based, and representativeness-based models. Uncertainty sampling \citep{Beluch_2018_CVPR, gorriz2017cost, wang2016cost,roy2001toward,ebrahimi2019uncertainty}, a type of heterogeneity-based model, is a standard active learning strategy where the learner aims to label those samples which have the most uncertain labelings. Non-Bayesian approaches\citep{brinker2003incorporating,wang2015querying} dealing with entropy, distance from decision boundary, etc., also exist but are not scalable for deep learning \cite{sener2017active}. Representation-based methods that aim at increasing the diversity in a batch\citep{jain2016active} have also been studied. However, most of these works have been studied in the context of classification or low-dimensional regression problems, and the literature on dense regression is still sparse. 

\vspace{-5pt}
\paragraph{Statistical Image models.} The $n \times n$ RGB images occupy the space of $\mathbb{R}^{3n^2}$. However, the structured images occupy a small region in that space. The statistical properties of the samples in this small structured space can be leveraged to generate synthetic data that have similar statistics to real-world structured images. 
For instance, the observation that natural images follow a power law with respect to the magnitude of their Fourier Transform (FT) formed the basis for Wiener image denoising\citep{simoncelli20054}, Dead Leaves models \citep{lee2001occlusion} and fractals as image models \citep{redies2008fractal,kataoka2020pre}. 
Similarly, works like~\citep{field1987relations,simoncelli20054,kretzmer1952statistics} showed that outputs of zero mean wavelets to natural images are sparse and follow a generalized Laplacian distribution. Works like~\citep{heeger1995pyramid, portilla2000parametric} showed statistical models capable of producing realistic-looking textures. The recent work~\citep{noise} takes this research a step closer to realistic image generation by learning from procedural noise processes and using the generated samples for pre-training the neural networks. However, it is only applied to classification.

\vspace{-5pt}
\paragraph{Super-resolution.}  This consists of CNN-based methods to enhance the resolution of the image~\citep{ledig2017photo,wang2018esrgan,upadhyay2019robust,upadhyay2019mixed}. Attention mechanism has proven to be ubiquitous, with \citep{woo2018cbam} introducing channel and spatial attention modules for adaptive feature refinement. Transformers-based endeavors such as \citep{Liang_2021_ICCV}, achieve state-of-the-art results using multi-head self-attention for SR. \citep{saharia2022image} uses a probabilistic diffusion model and performs SR through an iterative denoising process. Works like~\citep{shocher2018zero,bose2022zero} use internal and external recurrence of information to get superior SR performance during inference. However, these works do not consider the problem of super-resolution in the active learning context, leaving a gap in the literature.

\vspace{-5pt}
\paragraph{Uncertainty Estimation.} Quantifying uncertainty in machine learning models is crucial for safety-critical applications~\citep{nair2020exploring,sudarshan2021towards,upadhyay2021uncertainty,upadhyayiccv,upadhyay2021robustness}. Uncertainty can be broadly categorized into two classes: (i) Epistemic uncertainty (i.e., uncertainty in model weights~\citep{blundell2015weight,daxberger2021laplace, graves2011practical,NIPS2017_2650d608}). (ii) Aleatoric uncertainty (i.e., noise inherent in the observations)~\citep{bae2021estimating,wang2019aleatoric}. 
The dense predictive uncertainty may be considered as a proxy for error and can be used for active learning purposes~\citep{laves2020well}.

\section{Method}
\label{sec:method}
\vspace{-5pt}
We first formulate the problem in Section \ref{sec:prob_form}, and present preliminaries on active learning, statistical image models, and uncertainty estimation in Section \ref{sec:pre_usim}. 
In Section~\ref{sec:usim}, we describe the construction of \textit{USIM-DAL} that learns a prior via statistical image modeling, which is later used to select the most informative samples from the unlabeled set for labeling and further improving the model.

\subsection{Problem formulation} 
\label{sec:prob_form}
\vspace{-5pt}

Let $\mathcal{D}_U = \{\mathbf{x}_i\}_{i=1}^{N}$ be the unlabeled set of input images from domain $\mathbf{X}$ (i.e., $\mathbf{x}_i \in \mathbf{X} \forall i$). 
We consider the task where images ($\mathbf{x}$)  are to be mapped to another set of dense continuous labels ($\mathbf{y}$, e.g., other images, such that $\mathbf{y}_i \in \mathbf{Y} \forall i$). We want to learn a mapping $\mathbf{\Psi}$ for the same, i.e.,
$\mathbf{\Psi}: \mathbf{X} \rightarrow \mathbf{Y}$.
However, we want to learn it under the constraint that we do not have sufficient \textit{budget} to ``label'' all the $N$ samples in $\mathcal{D}_U $ (i.e., acquire all the corresponding $\mathbf{y}$), but we do have a budget to label a significantly smaller subset of $\mathcal{D}_U$ with $K << N$ samples, say $\mathcal{D}_U^K$. This is a real-world constraint, as discussed in Section~\ref{sec:rw}. In this work, we focus on the problem of super-resolution where the domain $\mathbf{Y}$ consists of high-resolution images (corresponding to the low-resolution images in domain $\mathbf{X}$).

We tackle the problem of choosing the set of $K << N$ samples ($\mathcal{D}_U^K$) that are highly representative of the entire unlabeled training set $\mathcal{D}_U$, such that the learned mapping $\mathbf{\Psi}$ on unseen data from a similar domain performs well.

\subsection{Preliminaries}
\label{sec:pre_usim}
\vspace{-5pt}

\paragraph{Active Learning.} 
As discussed above, given a set of $N$ unlabeled images $\mathcal{D}_U$, we want to choose a set of $K << N$ samples ($\mathcal{D}_U^K$) that are highly representative of the entire unlabeled training set $\mathcal{D}_U$.
This is the problem of active learning, which consists of \textit{query strategies} that maps the entire unlabeled set $\mathcal{D}_U$ to its subset. That is, the query strategy (constrained to choose $K$ samples and parameterized by $\phi$) is given by,
$
    \mathcal{Q}_{K,\phi}: \mathcal{D}_U \rightarrow \mathcal{D}_U^K
$.
Many works explore designing the query strategy $\mathcal{Q}_{K,\phi}$~\citep{Beluch_2018_CVPR, gorriz2017cost, wang2016cost}. However, they seldom attempt to design such a strategy for dense regression.

\begin{figure}[!h]
    \centering
    \includegraphics[width=0.49\textwidth]{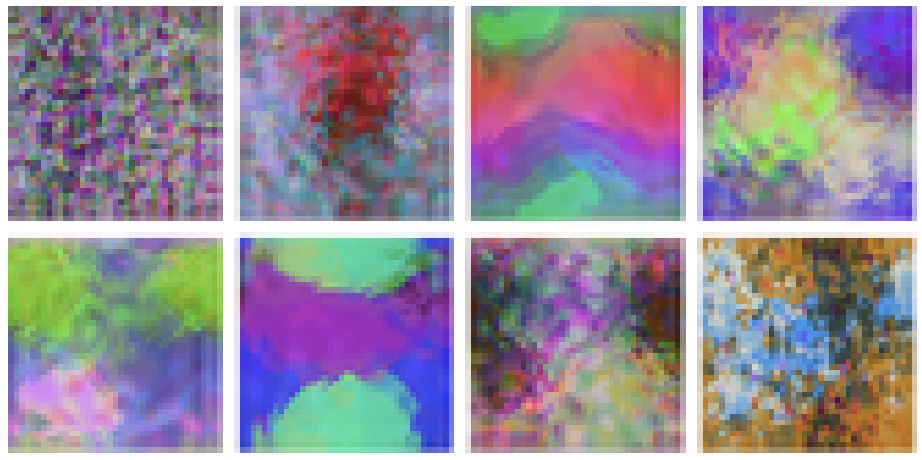}
    \vspace{-10pt}
    \caption{Samples generated from Statistical Image Models (combination of Spectrum + WMM + Color histogram).}
    \vspace{-10pt}
    \label{fig:sim}
\end{figure}
\paragraph{Statistical Image Models (SIM).} 
As discussed in~\citep{noise}, the statistical properties of RGB images can be exploited to generate synthetic images that can serve as an excellent pre-training learning signal. The generative model (based on statistical properties of RGB images) is described as $\mathcal{G}(\cdot; \theta_G): \mathbf{z} \rightarrow \mathbf{x}$ where $\mathbf{z}$ is a stochastic latent variable and $\mathbf{x}$ is an image. The image generation is modelled as a hierarchical process in which, first, the parameters of a model are sampled. Then the image is sampled given these parameters and stochastic noise. Previous works~\citep{noise} highlight the following statistical models.
(i)~\textbf{Spectrum:} based on the magnitude of the Fourier transform (FT). The FT of many natural images follows a power law, i.e.,  $\frac{1}{|f|^{\alpha}}$, where $|f|$ is the magnitude of frequency $f$, and $\alpha$ is a constant close to 1. For generative models, the sampled images are constrained to be random noise images that have FT magnitude following $\frac{1}{|f_x|^a + |f_y|^b}$ with a and b being two random numbers uniformly sampled as detailed in~\citep{noise}.
(ii)~\textbf{Wavelet-marginal model (WMM):} Generates the texture by modeling their histograms of wavelet coefficient as discussed in~\citep{simoncelli20054,kretzmer1952statistics}.
(iii)~\textbf{Color histograms:} As discussed in~\citep{noise}, this generative model follows the color distribution of the dead-leaves model~\citep{noise}.
Combining all these different models allows for capturing colour distributions, spectral components, and wavelet distributions that mimic those typical for natural images. 
Figure~\ref{fig:sim} shows examples of generated samples from such models.

\paragraph{Uncertainty Estimation.} 
Various works~\citep{lakshminarayanan2016simple,NIPS2017_2650d608} have proposed different methods to model the uncertainty estimates in the predictions made by DNNs for different tasks.
Interestingly recent works~\citep{NIPS2017_2650d608,upadhyay2022bayescap} have shown that for many real-world vision applications, modeling the aleatoric uncertainty allows for capturing erroneous predictions that may happen with out-of-distribution samples. 
To estimate the uncertainty for the regression tasks using deep network (say $\mathbf{\Psi}(\cdot; \zeta): \mathbf{X} \rightarrow \mathbf{Y}$), the model must capture the output distribution $\mathcal{P}_{Y|X}$. This is often done by estimating $\mathcal{P}_{Y|X}$ with a parametric distribution and learning the parameters of the said distribution using the deep network, which is then used to maximize the likelihood function. 
That is,
for an input $\mathbf{x}_i$, the model produces a set of parameters representing the output given by, $\{\hat{\mathbf{y}}_i, \hat{\nu}_i \dots \hat{\rho}_i \} := \mathbf{\Psi}(\mathbf{x}_i; \zeta)$, that characterizes the distribution 
$\mathcal{P}_{Y|X}(\mathbf{y}; \{\hat{\mathbf{y}}_i, \hat{\nu}_i \dots \hat{\rho}_i \})$, such that 
$\mathbf{y}_i \sim \mathcal{P}_{Y|X}(\mathbf{y}; \{\hat{\mathbf{y}}_i, \hat{\nu}_i \dots \hat{\rho}_i \})$.
The likelihood $\mathscr{L}(\zeta; \mathcal{D}) := \prod_{i=1}^{N} \mathcal{P}_{Y|X}(\mathbf{y}_i; \{\hat{\mathbf{y}}_i, \hat{\nu}_i \dots \hat{\rho}_i \})$ is then maximized to estimate the optimal parameters of the network.
Typically, the parameterized distribution is chosen to be \textit{heteroscedastic} Gaussian distribution, in which case $\mathbf{\Psi}(\cdot; \zeta)$ is designed to predict the \textit{mean} and \textit{variance} of the Gaussian distribution, i.e., 
$\{\hat{\mathbf{y}}_i, \hat{\sigma}_i^2 \} := \mathbf{\Psi}(\mathbf{x}_i; \zeta)$. The optimization problem becomes,
\begin{gather}
\zeta^* =  
\underset{\zeta}{\text{argmin}} \sum_{i=1}^{N} \frac{|\hat{\mathbf{y}}_i - \mathbf{y}_i|^2}{2\hat{\sigma}_i^2} + \frac{\log(\hat{\sigma}_i^2)}{2} \label{eq:unc}
\end{gather}
With $\text{Uncertainty}(\hat{\mathbf{y}}_i) = \hat{\sigma}_i^2$.
An important observation from Equation~\ref{eq:unc} is that,
ignoring the dependence through $\zeta$, the solution to Equation~\ref{eq:unc} decouples estimation of $\hat{\mathbf{y}}_i$ and $\hat{\mathbf{\sigma}}_i$. That is, for minimizing with respect to $\hat{\mathbf{y}}_i$ we need, 
\begin{gather}
    \frac{\partial \left( \sum_{i=1}^{N} \frac{|\hat{\mathbf{y}}_i - \mathbf{y}_i|^2}{2\hat{\sigma}_i^2} + \frac{\log(\hat{\sigma}_i^2)}{2} \right)}
    {\partial \hat{\mathbf{y}}_i} = 0 \label{eq:unc_d1} \\
    \frac{\partial^2 \left( \sum_{i=1}^{N} \frac{|\hat{\mathbf{y}}_i - \mathbf{y}_i|^2}{2\hat{\sigma}_i^2} + \frac{\log(\hat{\sigma}_i^2)}{2} \right)}
    {\partial \hat{\mathbf{y}}_i^2} > 0 \label{eq:unc_d2}
\end{gather}
Equation~\ref{eq:unc_d1} \& \ref{eq:unc_d2} lead to $\hat{\mathbf{y}}_i = \mathbf{y}_i \text{ }\forall i$. Similarly for minimizing with respect to $\hat{\sigma}_i$ we need,
\begin{gather}
    \frac{\partial \left( \sum_{i=1}^{N} \frac{|\hat{\mathbf{y}}_i - \mathbf{y}_i|^2}{2\hat{\sigma}_i^2} + \frac{\log(\hat{\sigma}_i^2)}{2} \right)}
    {\partial \hat{\mathbf{\sigma}}_i} = 0 \label{eq:unc_d3} \\
    \frac{\partial^2 \left( \sum_{i=1}^{N} \frac{|\hat{\mathbf{y}}_i - \mathbf{y}_i|^2}{2\hat{\sigma}_i^2} + \frac{\log(\hat{\sigma}_i^2)}{2} \right)}
    {\partial \hat{\mathbf{\sigma}}_i^2} > 0 \label{eq:unc_d4}
\end{gather}
Equation~\ref{eq:unc_d3} \& \ref{eq:unc_d4} lead to $\hat{\mathbf{\sigma}}_i^2 = |\hat{\mathbf{y}_i} - \mathbf{y}_i|^2 \text{ }\forall i$.
That is, the estimation $\hat{\mathbf{\sigma}}_i^2$ should perfectly reflect the squared error. Therefore, a higher $\hat{\mathbf{\sigma}}_i^2$ indicates higher error. We leverage this observation to design our dense active learning framework as described in Section~\ref{sec:usim}.

\begin{figure*}[!t]
    \centering
    \includegraphics[width=\textwidth]{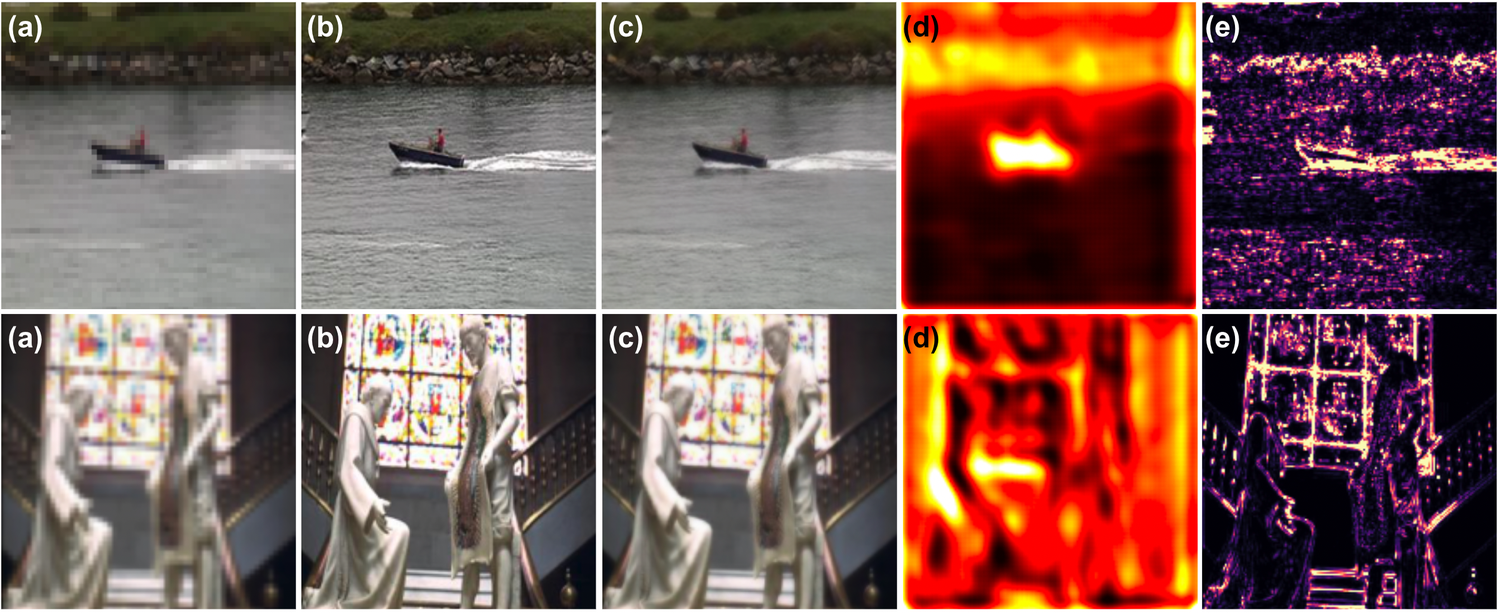}
    \vspace{-15pt}
    \caption{Output of the pre-trained probabilistic deep network (which is trained using synthetic images sampled from statistical image models) on samples from \textit{unseen} natural image datasets. (a)~LR input, (b)~HR groundtruth, (c)~Predicted output, SR, from the network, (d)~Predicted uncertainty from the network, (e)~Error between SR and groundtruth.}
    \label{fig:dal_u}
    \vspace{-10pt}
\end{figure*}

\vspace{-10pt}
\subsection{Constructing \textit{USIM-DAL}}
\label{sec:usim}
\vspace{-5pt}
To tackle the problem mentioned in Section~\ref{sec:prob_form} (i.e., choosing a small subset), we leverage the fact that even before training the model with the labelled set, we can train a model based on the samples that we get from statistical image model as described above, which can then be used to make inference on the unlabeled domain-specific dataset identifying the high-uncertainty samples. The high-uncertainty samples can then be labelled and used to fine-tune the model.

We constraint the generative process for statistical image models as,
Similar to~\citep{noise}, we treat image generation as a hierarchical process in which first the parameters of a model, $\theta_G$, are sampled. Then the image is sampled given these parameters and stochastic noise, i.e.,
\begin{gather}
    \theta_G \sim prior(\theta_G)  \text{ and } \mathbf{z} \sim prior(\mathbf{z}) \\
    \mathbf{x} = \mathcal{G}(\mathbf{z}; \theta_G)
\end{gather}
In particular, for super-resolution, we create a large (synthetic) labelled dataset using the samples from the statistical image models, say $\mathcal{D}_{SL} = \{(\text{\texttt{low}}(\mathbf{x}_{s,i}), \mathbf{x}_{s,i})\}_{i=1}^{M}$. Where $\mathbf{x}_{s,i}$ are generated samples from statistical image model and \texttt{low}$(\cdot)$, is the 4$\times$ down-sampling operation. We then train the network $\mathbf{\Psi}(\cdot; \zeta)$ on $\mathcal{D}_{SL}$ using Equation~\ref{eq:unc}, leading to the optimal parameter $\zeta_{SL}^*$, as shown in Figure~\ref{fig:arch}. 
The trained model $\mathbf{\Psi}(\cdot; \zeta_{SL}^*)$ is then run in inference mode on all the samples of the unlabeled set $\mathcal{D}_{U}$ and gather the top uncertain samples for labeling, that is,
\begin{gather}
    \{ \hat{\mathbf{y}}_i, \hat{\mathbf{\sigma}}_i \} := \mathbf{\Psi}(\mathbf{x}_i; \zeta_{SL}^*) \text{ } \forall \mathbf{x}_i \in \mathcal{D}_{U} \\
    \mathcal{D}_{U}^{K} := \{ \mathbf{x}_j \} \forall j \in \text{\texttt{topK}}\left(\{ 
 \left< \hat{\mathbf{\sigma}}_i \right> \}_{i=1}^{N} \right)
\end{gather}
Where, $\left< \cdot \right>$ represents the mean operation, and $\text{\texttt{topK}}\left(\{ 
 \left< \hat{\mathbf{\sigma}}_i \right> \}_{i=1}^{N} \right)$ returns the indices of ``top-K'' most uncertain samples (i.e., mean uncertainty is high). We then acquire the labels for the samples in $\mathcal{D}_{U}^{K}$, giving us, $\mathcal{D}_{UL}^{K} = \{ (\mathbf{x}_j, \mathbf{y}_j) \}$.
 As discussed in Section~\ref{sec:pre_usim}, the input samples in $\mathbf{D}_{UL}^K$ serve as a proxy to the set of $K$ samples that would have the highest error between the prediction made by the model $\mathbf{\Psi}(\cdot; \zeta_{SL}^*)$ and the ground truth. That leads to better fine-tuning.
 The model $\mathbf{\Psi}(\cdot; \zeta_{SL}^*)$ is then fine-tuned on $\mathcal{D}_{UL}^{K}$ via Equation~\ref{eq:unc}, leading to the final state of the model $\mathbf{\Psi}(\cdot; \zeta_{KL}^*)$ (shown in Figure~\ref{fig:arch}) that can be used for inferring on the new sample.

 \textit{USIM-DAL} models the aleatoric uncertainties in the prediction. Still, it is crucial to note that it leverages the Statistical Image Modeling (SIM)-based synthetic images for pertaining and learning important priors for color images that broadly capture different niche domains such as medical images, satellite images, etc. Therefore, the initial model, capable of estimating the aleatoric uncertainty (trained on SIM-based synthetic images), can reasonably capture the uncertainty as a proxy for reconstruction error for domain-specific images that are not necessarily out-of-distribution images. Moreover, picking samples with high reconstruction errors for subsequent fine-tuning of the model yields better performance on similar highly erroneous cases, iteratively improving the model. Furthermore, in high-dimensional regression cases, the aleatoric and epistemic uncertainty often influence each other and are not independent~\cite{NIPS2017_2650d608,upadhyay2022bayescap,zhang2019reducing}. 

\vspace{-10pt}
\section{Experiments and Results}
\label{sec:exp}
\vspace{-5pt}

We provide an overview of the experiments performed and the results obtained. In Section~\ref{sec:comp}, we describe the task and various methods used for comparison. Section~\ref{sec:usim_exp} analyzes the performance of various dense active learning algorithms for super-resolution and shows that our proposed method \textit{USIM-DAL} can help greatly improve the performance when constrained with a limited budget.

\vspace{-10pt}
\subsection{Tasks, Datasets, and Methods}
\label{sec:comp}
\vspace{-5pt}
We present the results of all our experiments on the super-resolution task.
We demonstrate our proposed framework using a probabilistic SRGAN (which is the adaptation of SRGAN~\citep{ledig2017photo} that estimates pixel-wise uncertainty as described in~\citep{NIPS2017_2650d608}) model.
We evaluate the performance of various models on a wide variety of domains like 
(i)~Natural Images (with Set5, Set14, BSD100, and Visual Genome dataset~\citep{ledig2017photo, 937655,krishna2017visual}).
(ii)~Satellite Images (with PatternNet dataset~\citep{zhou2018patternnet}).
(ii)~Histopathology Medical Images (with Camelyon dataset~\citep{litjens20181399}).
The evaluation protocol is designed to constraint all the training domain datasets to be restricted by a small fixed number of images (also called \textit{training budget}). We used different training budgets of 500, 1000, 2000, 3000 and 5000 images for natural and satellite domains.
For both natural and satellite images, the input image resolution was set to $64\times64$. For natural images the training dataset was obtained from Visual Genome (separate from the test-set).
Similarly, for the histopathology medical images, the input image resolution was set to $32\times32$ and we used training budgets of 4000, 8000, 12000, and 16000.

We compare the super-resolution performance in terms of metrics MSE, MAE, PSNR, and SSIM~\citep{wang2004image} for the following methods on respective test sets:
(i)~SRGAN model trained from scratch with a randomly chosen subset satisfying the training budget from the entire training data (called \textit{Random}).
(ii)~SRGAN model trained from scratch on a large synthetically generated dataset via statistical image modeling (as described in Section~\ref{sec:pre_usim}). This model is called \textit{SIM}.
(iii)~SRGAN model trained from scratch on a large synthetically generated dataset via statistical image modeling and then fine-tuned on a randomly chosen subset satisfying the training budget from the entire training data, called \textit{SIM+Random}.
(iv)~SRGAN model trained from scratch on a large synthetically generated dataset via statistical image modeling and then fine-tuned on a subset chosen using uncertainty estimates, satisfying the training budget from the entire training data, called \textit{USIM-DAL}.

\vspace{-10pt}
\subsection{Dense Active Learning via Uncertainty Estimation}
\label{sec:dal_u}
\vspace{-5pt}
\begin{figure}[!h]
    \centering
    \includegraphics[width=0.45\textwidth]{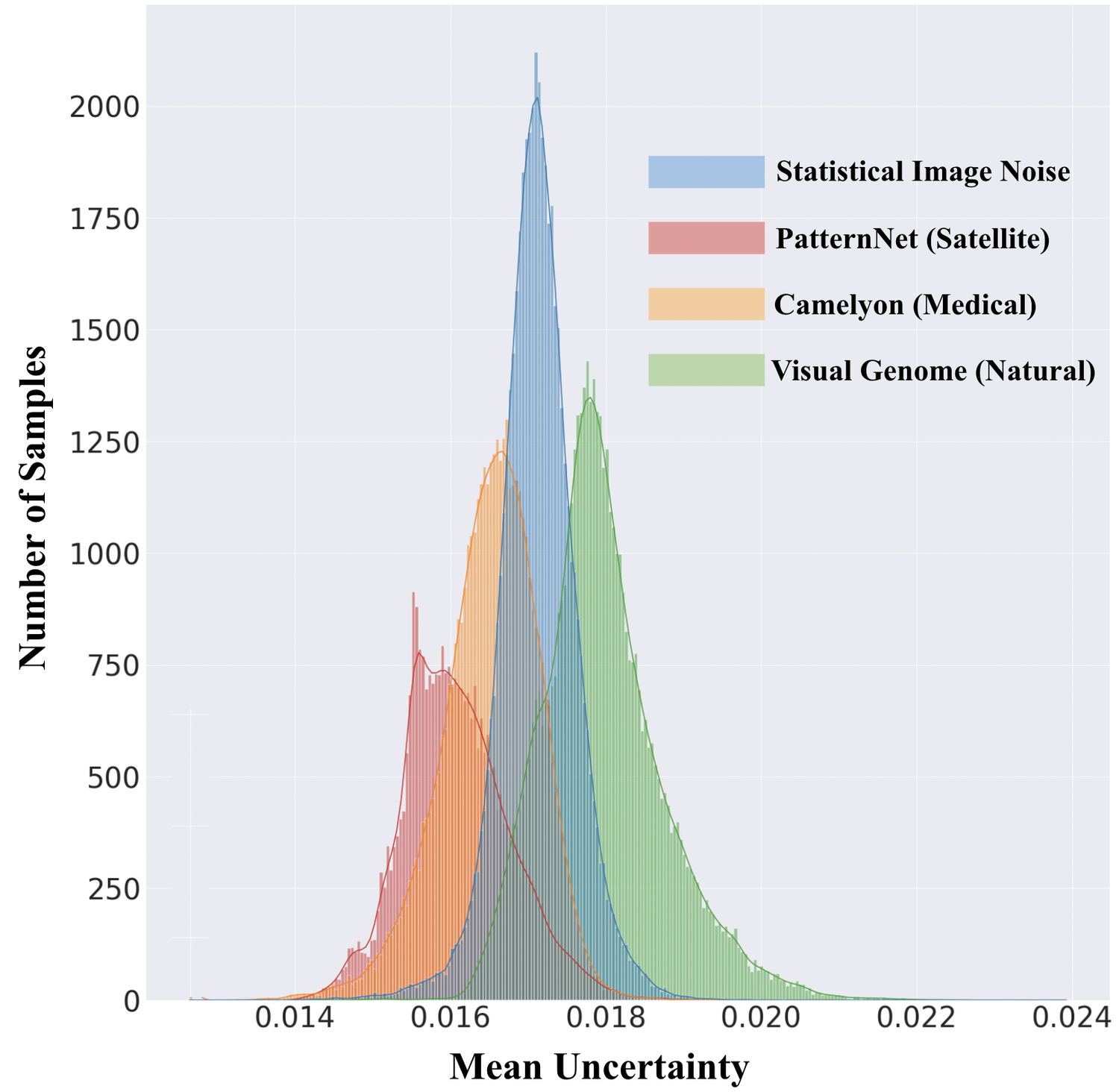}
    \vspace{-10pt}
    \caption{Distribution of mean uncertainty for samples in Statistical Image Noise, PatternNet (satellite), Camelyon (medical), Visual Genome (natural) datasets.}
    \label{fig:unc_dist}
    \vspace{-10pt}
\end{figure}
Our method proposes to utilize a probabilistic network that is learned from synthetic images sampled from statistical image models (i.e., $\mathbf{\Psi}(\cdot; \zeta^*_{SL})$ mentioned in Section~\ref{sec:usim}). 
Figure~\ref{fig:dal_u} shows the output of probabilistic SRGAN trained on synthetic images evaluated on samples from natural images. We observe that
(i)~The predicted super-resolved images (Figure~\ref{fig:dal_u}-(c)) are still reasonable.
(ii)~The uncertainty estimates (Figure~\ref{fig:dal_u}-(d)) still resemble the structures from the images and are a reasonable proxy to the error maps (Figure~\ref{fig:dal_u}-(e)) between the predictions and the ground truth, even though the model has never seen the natural images. 

\begin{figure}[!h]
    \centering
    \includegraphics[width=0.49\textwidth]{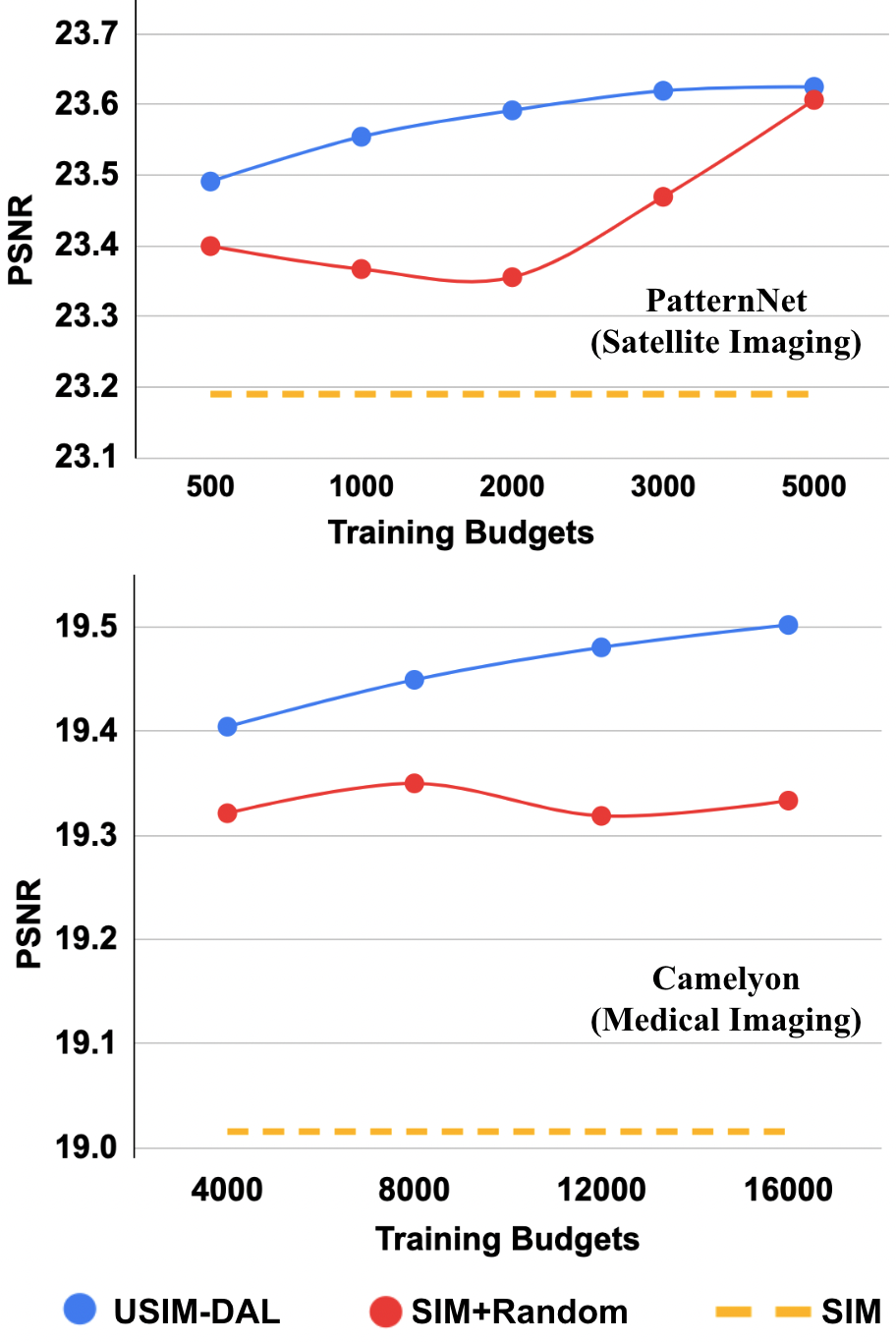}
    \vspace{-10pt}
    \caption{Evaluation of various methods on histopathology medical domain (i.e., Camelyon dataset) and satellite imaging domain (i.e., PatternNet dataset) at various fine-tuning budgets. The yellow curve is the \textit{SIM} baseline. The red curve is the SIM model fine-tuned with random samples (i.e., \textit{SIM+Random}). The blue curve is the SIM model fine-tuned with the highest uncertain samples (i.e., \textit{USIM-DAL}). }
    \label{fig:satmed}
    \vspace{-5pt}
\end{figure}

{
\setlength{\tabcolsep}{10pt}
\renewcommand{\arraystretch}{2.5}
\begin{table*}[!t]
\resizebox{\linewidth}{!}{
\begin{tabular}{c|c|ccccc}
  \multirow{4}{*}{\textbf{D}} & \multirow{4}{*}{\textbf{Methods}} & \multicolumn{5}{c}{\textbf{Budgets (Number of images)}} \\
  & & \textbf{500} & \textbf{1000} & \textbf{2000} & \textbf{3000} & \textbf{5000} \\
  & & \textbf{$\underbrace{\text{MSE}}_{\times 10^{3}}$} / \textbf{$\underbrace{\text{MAE}}_{\times 10^{2}}$} / \textbf{$\underbrace{\text{PSNR}}_{\times 10^{0}}$} / \textbf{$\underbrace{\text{SSIM}}_{\times 10^{2}}$} & \textbf{$\underbrace{\text{MSE}}_{\times 10^{3}}$} / \textbf{$\underbrace{\text{MAE}}_{\times 10^{2}}$} / \textbf{$\underbrace{\text{PSNR}}_{\times 10^{0}}$} / \textbf{$\underbrace{\text{SSIM}}_{\times 10^{2}}$} & \textbf{$\underbrace{\text{MSE}}_{\times 10^{3}}$} / \textbf{$\underbrace{\text{MAE}}_{\times 10^{2}}$} / \textbf{$\underbrace{\text{PSNR}}_{\times 10^{0}}$} / \textbf{$\underbrace{\text{SSIM}}_{\times 10^{2}}$} & \textbf{$\underbrace{\text{MSE}}_{\times 10^{3}}$} / \textbf{$\underbrace{\text{MAE}}_{\times 10^{2}}$} / \textbf{$\underbrace{\text{PSNR}}_{\times 10^{0}}$} / \textbf{$\underbrace{\text{SSIM}}_{\times 10^{2}}$} & \textbf{$\underbrace{\text{MSE}}_{\times 10^{3}}$} / \textbf{$\underbrace{\text{MAE}}_{\times 10^{2}}$} / \textbf{$\underbrace{\text{PSNR}}_{\times 10^{0}}$} / \textbf{$\underbrace{\text{SSIM}}_{\times 10^{2}}$} \\
  & & & & & & \\
   \hline

  \multirow{4}{*}{\rotatebox[origin=c]{90}{Set5}}
  & Random & 4.129 / 3.854 / 24.784 / 7.232 & 3.898 / 3.720 / 24.957 / 7.319 & 3.660 / 3.588 / 25.271 / 7.422 & 3.586 / 3.529 / 25.334 / 7.465 & 3.500 / 3.420 / 25.514 / 7.539 \\
  & SIM & 3.431 / 3.524 / 25.641 / 7.541 & 3.431 / 3.524 / 25.641 / 7.541 & 3.431 / 3.524 / 25.641 / 7.541 & 3.431 / 3.524 / 25.641 / 7.541 & 3.431 / 3.524 / 25.641 / 7.541 \\
  & SIM + Random & 2.976 / 3.139 / 26.283 / 7.839 & 2.958 / 3.099 / 26.377 / 7.872 & 2.941 / 3.081 / 26.435 / 7.896 & 2.934 / 3.088 / 26.436 / 7.910 & 2.912 / 3.056 / 26.546 / 7.935 \\
  & USIM-DAL & \textbf{2.926} / \textbf{3.088} / \textbf{26.484} / \textbf{7.869} & \textbf{2.884} / \textbf{3.069} / \textbf{26.550} / \textbf{7.894} & \textbf{2.848} / \textbf{3.027} / \textbf{26.619} / \textbf{7.931} & \textbf{2.843} / \textbf{3.029} / \textbf{26.644} / \textbf{7.944} & \textbf{2.831} / \textbf{3.025} / \textbf{26.699} / \textbf{7.943} \\
  \hline
  \hline
  \multirow{4}{*}{\rotatebox[origin=c]{90}{Set14}}
  & Random & 6.254 / 4.750 / 22.535 / 6.333 & 6.111 / 4.669 / 22.576 / 6.382 & 5.942 / 4.564 / 22.701 / 6.468 & 5.862 / 4.539 / 22.616 / 6.488 & 5.800 / 4.450 / 22.886 / 5.594 \\
  & SIM & 4.852 / 4.303 / 22.897 / 6.383 & 4.852 / 4.303 / 22.897 / 6.383 & 4.852 / 4.303 / 22.897 / 6.383 & 4.852 / 4.303 / 22.897 / 6.383 & 4.852 / 4.303 / 22.897 / 6.383 \\
  & SIM + Random & 4.488 / 3.907 / 23.748 / \textbf{7.016} & 4.485 / 3.871 / 23.787 / \textbf{7.082} & 4.444 / 3.828 / 24.106 / 7.159 & 4.426 / 3.828 / 24.162 / 7.179 & 4.396 / 3.798 / 24.090 / 7.198 \\
  & USIM-DAL & \textbf{4.376} / \textbf{3.836} / \textbf{23.810} / 6.984 & \textbf{4.366} / \textbf{3.816} / \textbf{23.818} / 7.000 & \textbf{4.331} / \textbf{3.767} / \textbf{24.288} / \textbf{7.177} & \textbf{4.317} / \textbf{3.749} / \textbf{24.422} / \textbf{7.208} & \textbf{4.292} / \textbf{3.728} / \textbf{24.553} / \textbf{7.227} \\
  \hline
  \hline
  \multirow{4}{*}{\rotatebox[origin=c]{90}{BSD100}}
  & Random & 4.857 / 4.338 / 23.357 / 6.072 & 4.778 / 4.294 / 23.427 / 6.098 & 4.670 / 4.226 / 23.583 / 6.160 & 4.630 / 4.207 / 23.598 / 6.187 & 4.600 / 4.160 / 23.703 / 6.214 \\
  & SIM & 3.526 / 3.738 / 24.805 / 6.713 & 3.526 / 3.738 / 24.805 / 6.713 & 3.526 / 3.738 / 24.805 / 6.713 & 3.526 / 3.738 / 24.805 / 6.713 & 3.526 / 3.738 / 24.805 / 6.713 \\
  & SIM + Random & 3.362 / 3.578 / 25.007 / 6.786 & 3.352 / 3.559 / 25.043 / 6.794 & 3.328 / 3.539 / 25.092 / 6.812 & 3.323 / 3.540 / 25.085 / 6.816 & 3.305 / 3.519 / 25.137 / 6.834 \\
  & USIM-DAL & \textbf{3.299} / \textbf{3.520} / \textbf{25.174} / \textbf{6.826} & \textbf{3.293} / \textbf{3.520} / \textbf{25.191} / \textbf{6.830} & \textbf{3.282} / \textbf{3.504} / \textbf{25.207} / \textbf{6.838} & \textbf{3.277} / \textbf{3.496} / \textbf{25.212} / \textbf{6.844} & \textbf{3.262} / \textbf{3.486} / \textbf{25.263} / \textbf{6.854} \\
  \hline
  \hline
  \multirow{4}{*}{\rotatebox[origin=c]{90}{Visual Genome}}
  & Random & 4.442 / 3.946 / 23.935 / 6.853 & 4.346 / 3.892 / 24.033 / 6.889 & 4.231 / 3.818 / 24.200 / 6.954 & 4.182 / 3.797 / 24.216 / 6.983 & 4.120 / 3.718 / 24.353 / 7.032 \\
  & SIM & 4.310 / 3.963 / 24.055 / 6.826 & 4.310 / 3.963 / 24.055 / 6.826 & 4.310 / 3.963 / 24.055 / 6.826 & 4.310 / 3.963 / 24.055 / 6.826 & 4.310 / 3.963 / 24.055 / 6.826 \\
  & SIM + Random & 4.038 / 3.721 / 24.396 / 7.036 & 4.026 / 3.690 / 24.423 / 7.056 & 3.993 / 3.663 / 24.496 / 7.088 & 3.977 / 3.661 / 24.515 / 7.101 & 3.943 / 3.631 / 24.563 / 7.126 \\
  & USIM-DAL & \textbf{3.966} / \textbf{3.668} / \textbf{24.543} / \textbf{7.056} & \textbf{3.949} / \textbf{3.657} / \textbf{24.570} / \textbf{7.069} & \textbf{3.925} / \textbf{3.623} / \textbf{24.624} / \textbf{7.109} & \textbf{3.908} / \textbf{3.608} / \textbf{24.656} / \textbf{7.126} & \textbf{3.880} / \textbf{3.593} / \textbf{24.721} / \textbf{7.143} \\
  \hline
  \hline
\end{tabular}%
}
\caption{
 Evaluating different methods on natural image datasets ( Set5, Set14, BSD100, Visual Genome) using MSE, MAE, PSNR, SSIM. Lower MSE/MAE is better. Higher PSNR/SSIM is better. ``D'': Datasets. Best results are in {\bf bold}. 
}
\vspace{-10pt}
\label{tab:t1}
\end{table*}
}

We use the predicted uncertainty from this model to identify the samples from the real-world domain that would lead to high errors. Figure~\ref{fig:unc_dist} shows the distribution of mean uncertainty values for samples in (i)~Statistical Noise (ii)~Natural (ii)~Satellite (iii)~Medical image datasets. We notice that the model trained on synthetic images leads to a gaussian distribution for the mean uncertainty values on the synthetic image datasets. We obtain similar distributions for other datasets from different domains. This further emphasizes that uncertainty estimates obtained from $\mathbf{\Psi}(\cdot; \zeta^*_{SL})$ can be used as a proxy to identify the highly uncertain (therefore erroneous) samples from different domains (i.e., the samples close to the right tail of the distributions).

\subsection{\textit{USIM-DAL} for Super-resolution}
\label{sec:usim_exp}
\vspace{-5pt}

Table~\ref{tab:t1} shows the performance of different methods on multiple natural image datasets, including Set5, Set14, BSD100, and Visual Genome (VG).
We observe that with the smallest training budget of 500 images, \textit{USIM-DAL} performs the best with a PSNR/MAE of 25.174/0.035 (Table~\ref{tab:t1} shows the results with a scaling factor for better accommodation) compared to \textit{SIM+Random} with PSNR/MAE of 25/0.039 and \textit{SIM} with PSNR/MAE of 24.8/0.037. We also notice that at this budget, choosing the random subset of the training dataset to train the model from scratch performs the worst with PSNR/MAE of 23.36/0.043.
As the budget increases (left to right in Tabel~\ref{tab:t1}), the performances of all the methods also improve. However, a similar trend is observed where the \textit{USIM-DAL} performs better than \textit{SIM+Random}, \textit{SIM}, and \textit{Random}.
We observe a similar trend for other natural image datasets.
This allows us to make the following observations:
(i)~Using a synthetic training image dataset (sampled from the statistical image model, discussed in Section~\ref{sec:pre_usim}) leads to better performance than using a small random subset of training images from the original domain (i.e., \textit{SIM} better than \textit{Random}).
(ii)~Using the above synthetic training image dataset to train a model and later fine-tuning it with domain-specific samples lead to further improvements (i.e., both \textit{USIM-DAL} and \textit{SIM+Random} better than \textit{SIM}).
(iii)~With a limited budget, fine-tuning a model (pre-trained on synthetic training image dataset) using high-uncertainty samples from the training set (as decided by the \textit{USIM-DAL}) is better than using the random samples from the training set (i.e., \textit{USIM-DAL} better than \textit{SIM+Random}).

We perform a similar set of experiments with other imaging domains, namely, (i)~Satellite imaging (using PatternNet dataset) and (ii)~Medical imaging (using Camelyon histopathology dataset). We observe a similar (to natural images) trend in these domains. 
Figure~\ref{fig:satmed} shows the performance (measured using PSNR) for different methods on these two domains, with varying training budgets. 
For satellite imaging, at the lowest training budget of 500 images, \textit{USIM-DAL} with PSNR of 23.5 performs better than \textit{SIM+Random} with PSNR of 23.4 and \textit{SIM} with a PSNR of 23.2. We observe that as the training budget increases to 2000 images, \textit{USIM-DAL} (with PSNR of 23.6) outperforms \textit{SIM+Random} (with PSNR of 23.35) with an even higher margin.
As we increase the training budget further, the \textit{SIM+Random} model starts performing similarly to \textit{USIM-DAL}. With a budget of 5000 samples, \textit{USIM-DAL} has a performance of 23.62, and \textit{SIM+Random} has a performance of 23.60. 
Given a domain with large (specific to datasets) training budgets, the performance achieved from random sampling and active learning strategies will converge.
\begin{figure}[!h]
    \centering
    \includegraphics[width=0.49\textwidth]{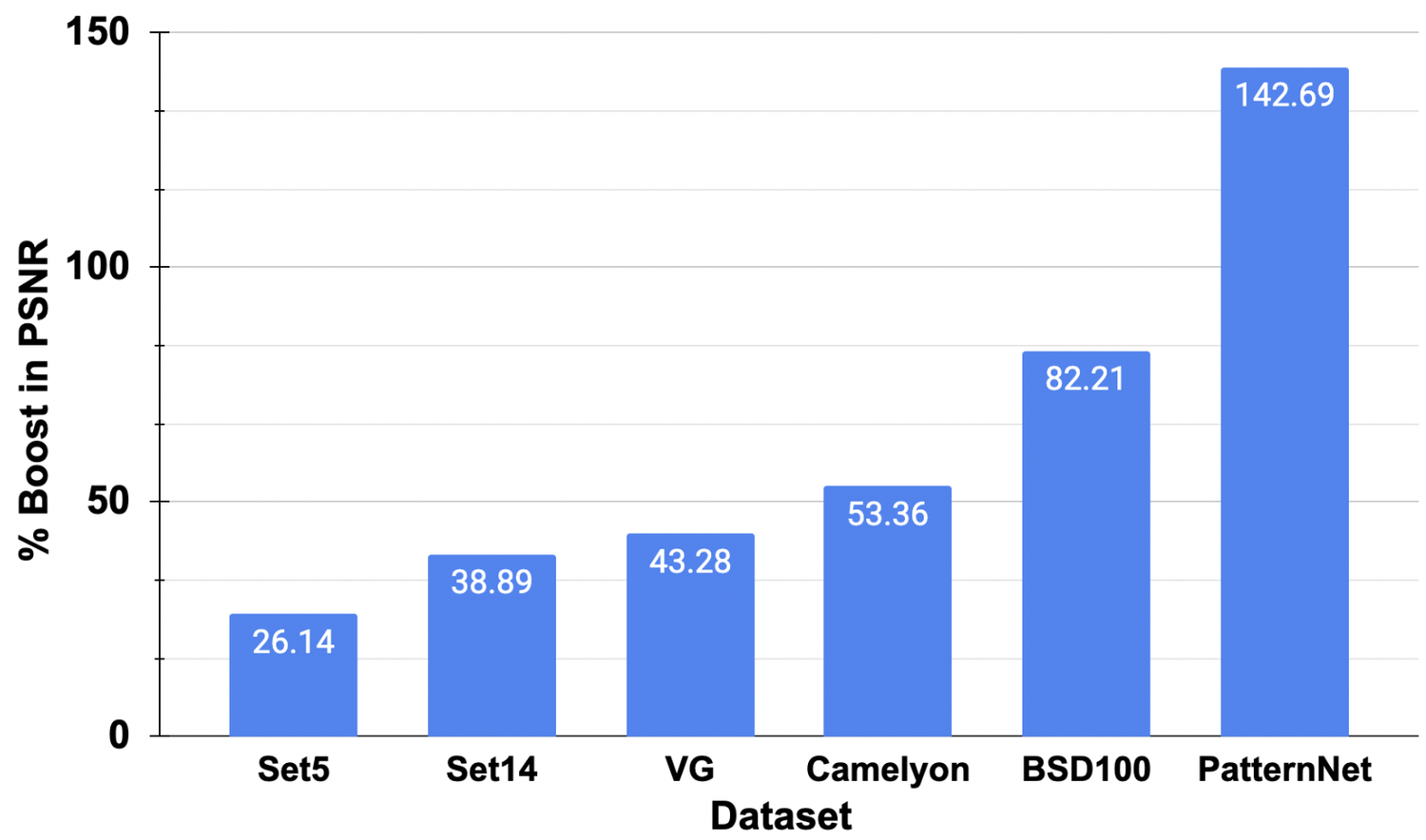}
    \vspace{-10pt}
    \caption{ Relative \% boost in PSNR of \textit{USIM-DAL} relative to \textit{SIM+Random} over \textit{SIM} baseline (Equation~\ref{eq:pboost}) at optimal budget for six datasets across three domains.  }
    \label{fig:pboost}
    \vspace{-10pt}
\end{figure}

\begin{figure*}[!t]
    \centering
    \includegraphics[width=0.49\textwidth]{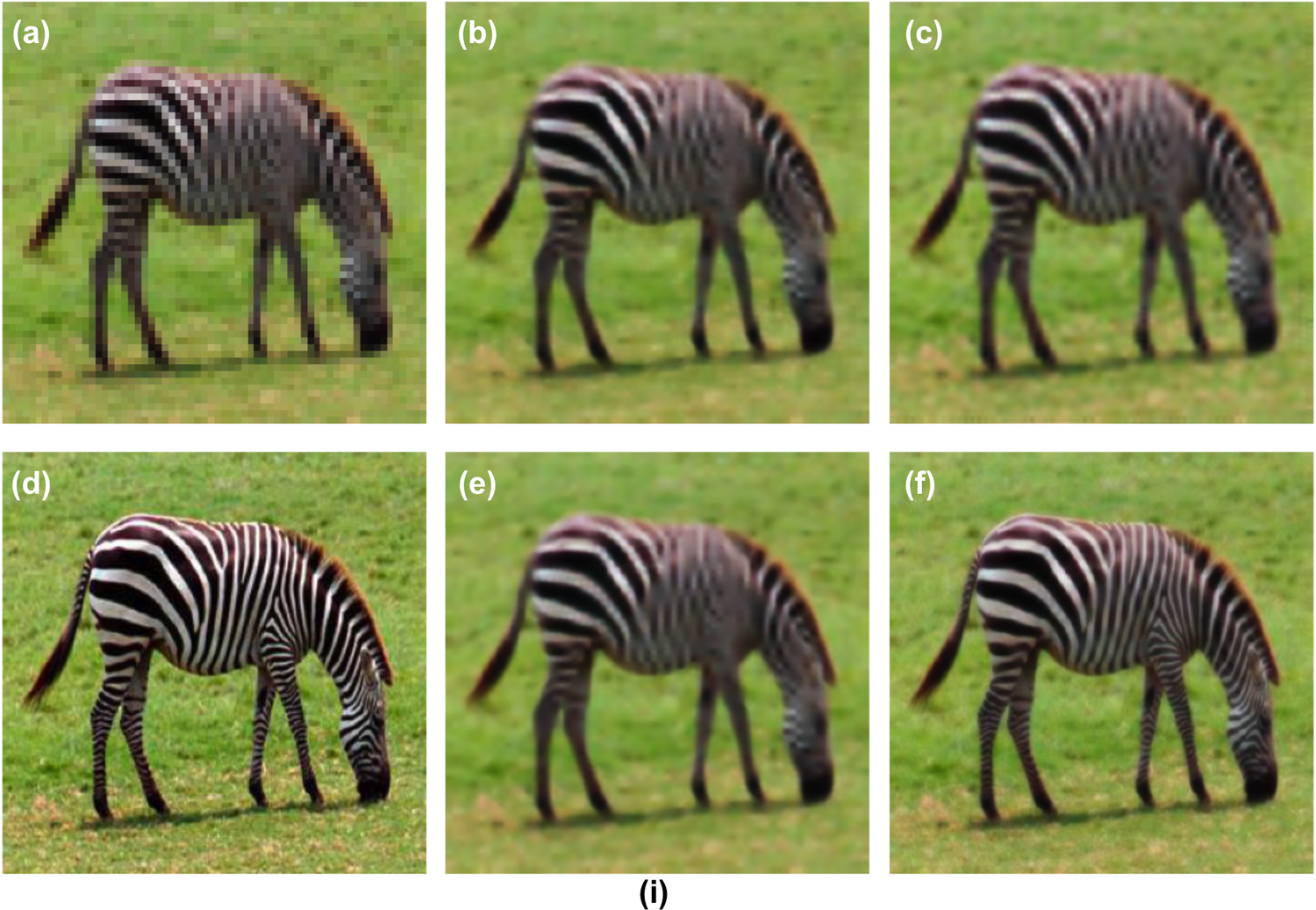}
    \includegraphics[width=0.49\textwidth]{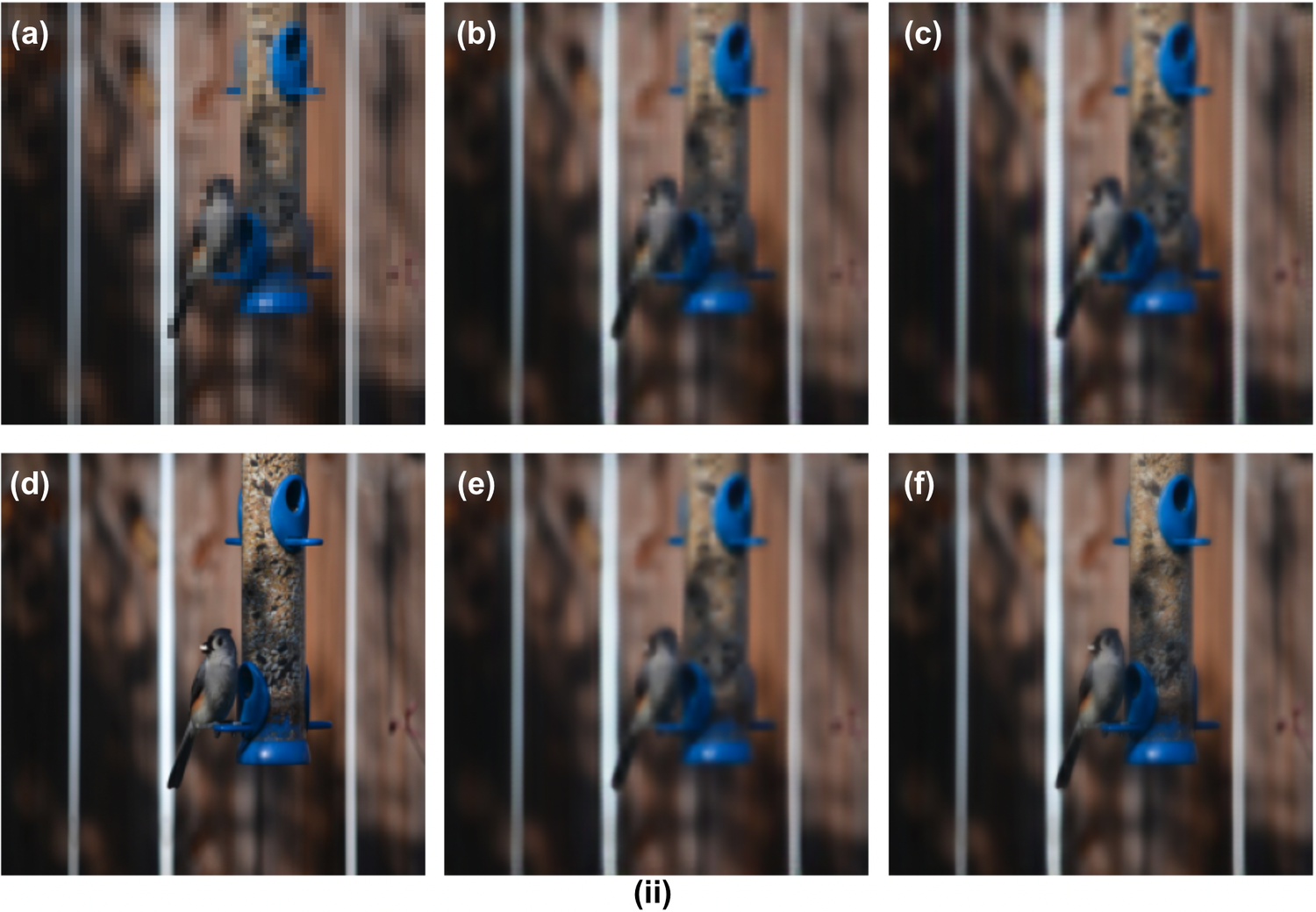}
    \includegraphics[width=0.49\textwidth]{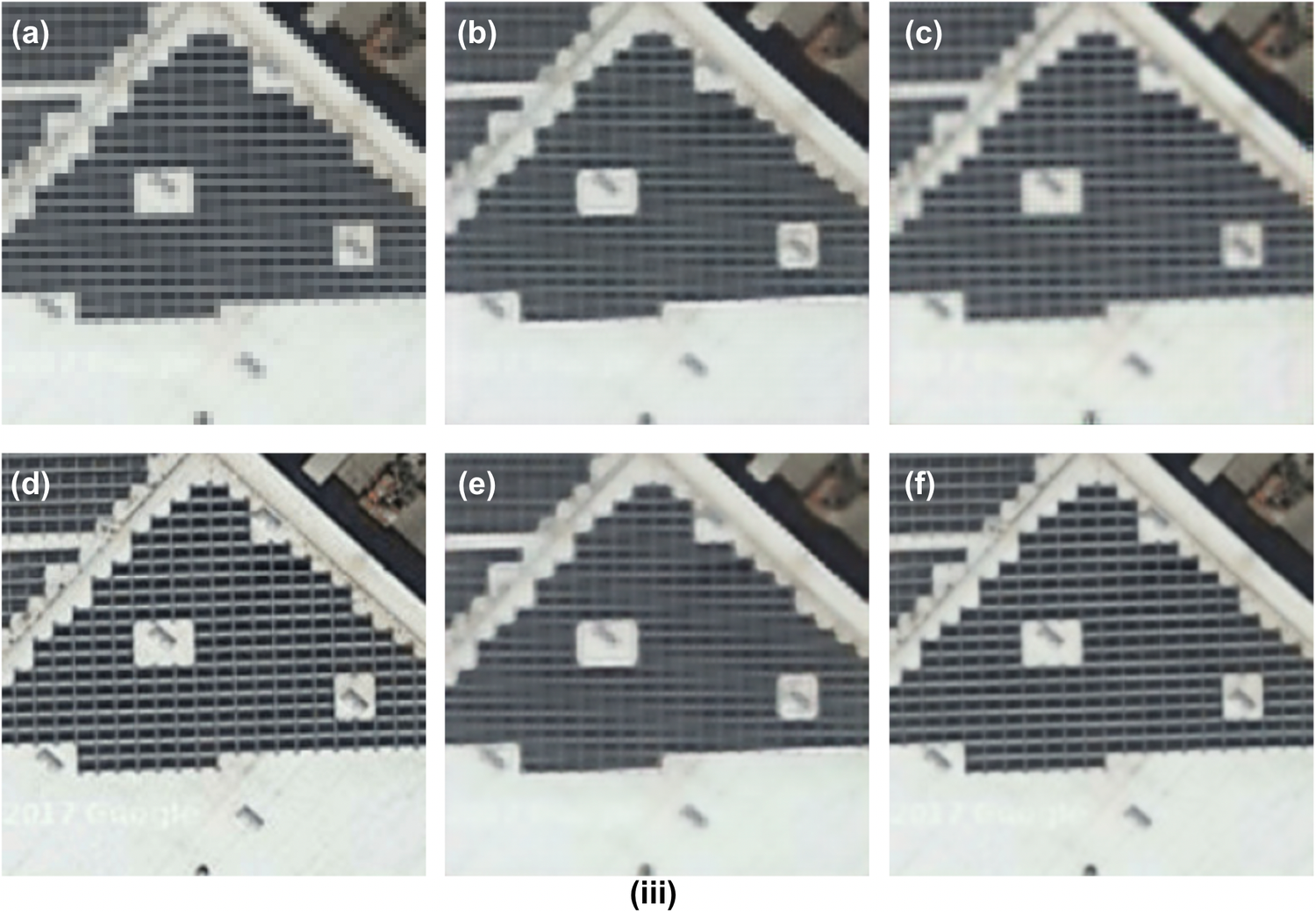}
    \includegraphics[width=0.49\textwidth]{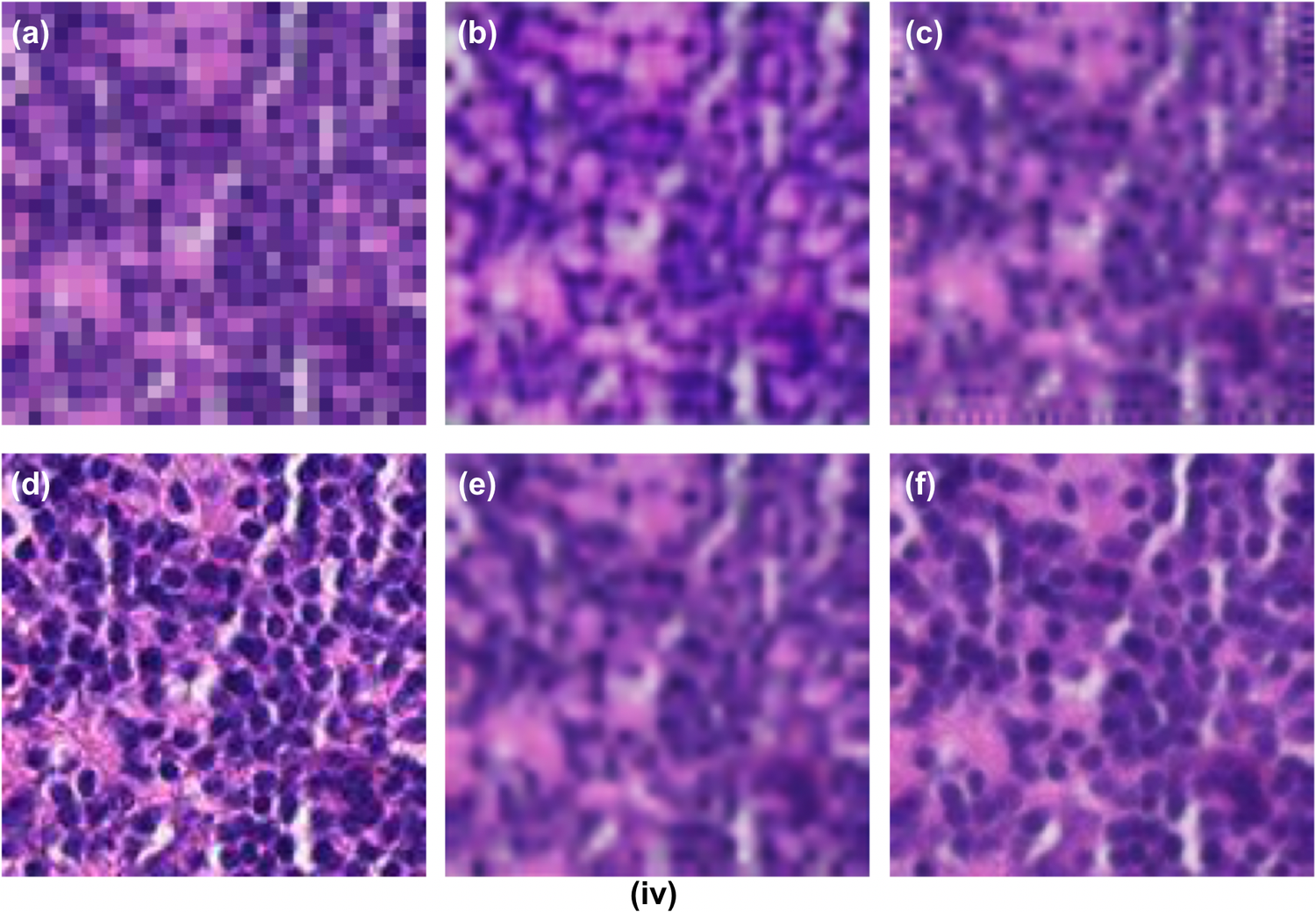}
    \caption{Qualitative results from different methods (performing 4$\times$ super-resolution) including (b)~\textit{Random}, (c)~\textit{SIM}, (e)~\textit{SIM+Random}, (f)~\textit{USIM-DAL} on (i)~BSD100, (ii)~Visual Genome, (iii)~PatternNet, and (iv)~Camelyon datasets. (a)~LR input, and (d)~HR groundtruth.
    Input resolution for BSD100, Visual Genome, and PatternNet is $64\times64$, and for Camelyon is $32\times32$.
    (f)~\textit{USIM-DAL} produces the most visually appealing outputs.
    }
    \label{fig:qual}
\end{figure*}

For Camelyon dataset, we use the input image resolution of 32$\times$32. We observe that \textit{USIM-DAL} performs the best across all budgets when compared to \textit{SIM+Random} and \textit{SIM}. We also note that high-frequency features that are typically present in high-resolution scans (i.e., obtained at 20$\times$ or 40$\times$ magnification from the histopathology microscope) make the super-resolution problem harder and require more data to achieve good performance.

Figure~\ref{fig:pboost} summarizes the performance gain (in terms of PSNR) by using \textit{USIM-DAL} (i.e., uncertainty-based active learning strategy for dense regression) compared to \textit{SIM+Random} (i.e., no active learning, randomly choosing a subset from real training domain), relative to \textit{SIM} (i.e., no real samples used from the domain) at best performing limited budgets. That is, the relative percentage boost in performance is reported as:
\begin{gather}
    \frac{(\text{PSNR}_{\text{USIM-DAL}} - \text{PSNR}_{\text{SIM+Random}})*100}{\text{PSNR}_{\text{SIM+Random}}-\text{PSNR}_{\text{SIM}}}
    \label{eq:pboost}
\end{gather}
We note that \textit{USIM-DAL} consistently performs better than \textit{SIM+Random}, with the relative percentage boost in PSNR of 26.14\% for Set5 to 142.69\% for PatternNet.
Figure~\ref{fig:qual} shows the qualitative outputs of different models on multiple datasets. 
On all the datasets, we notice that the output obtained by \textit{USIM-DAL} is better than the output of \textit{SIM+Random} that is better than \textit{SIM} and \textit{Random}.

\section{Discussion and Conclusion}
In this work, we presented a novel framework called \textit{USIM-DAL} that is designed to perform active learning for dense-regression tasks, such as image super-resolution. 
Dense-regression tasks, such as super-resolution, are an important class of problem for which deep learning offers a wide range of solutions applicable to medical imaging, security, and remote sensing. However, most of these solutions often rely on supervision signals derived from high-resolution images.
Due to the time-consuming acquisition of high-resolution images or expensive sensors, hardware, and operational costs involved, it is not always feasible to generate large volumes of high-resolution imaging data. But in real-world scenarios, a limited budget for acquiring high-resolution data is often available. This calls for active learning that chooses a subset from  large unlabeled set to perform labeling to train the models.
While multiple querying strategies (in the context of active learning) exist for the classification tasks, the same for dense regression tasks are seldom discussed. 
Our work paves the way for using modern uncertainty estimation techniques for active learning in dense regression tasks. 
We show that a large synthetic dataset acquired using statistical image models can be used to learn informative priors for various domains, including natural images, medical images, satellite images, and more.  
The learned prior can then be used to choose the subset consisting of high-uncertainty samples that can then be labeled and used to fine-tune the prior further.
Through extensive experimentation,
we show that our approach generalizes well to a wide variety of domains, including medical and satellite imaging.
we show that active learning performed by proposed querying strategy (i.e., \textit{USIM-DAL}) leads to gains of upto 140\% / 53\% with respect to a random selection strategy (i.e., SIM+Random) relative to no dataset-specific fine-tuning (i.e., \textit{SIM}) on satellite/medical imaging.

\textbf{Acknowledgements.} This work has been partially funded by the ERC (853489 - DEXIM) and by
the DFG (2064/1 – Project number 390727645). The authors thank the International Max Planck
Research School for Intelligent Systems (IMPRS-IS) for supporting Uddeshya Upadhyay.

\bibliography{rangnekar_644}
\end{document}